\documentclass[conference]{IEEEtran}
\IEEEoverridecommandlockouts
\usepackage{cite}
\usepackage{amsmath,amssymb,amsfonts}
\usepackage{algorithmic}
\usepackage{graphicx}
\usepackage{subfig}
\usepackage{txfonts}
\usepackage{dsfont}
\usepackage{caption}

\usepackage[ruled,linesnumbered]{algorithm2e}

\def\BibTeX{{\rm B\kern-.05em{\sc i\kern-.025em b}\kern-.08em
    T\kern-.1667em\lower.7ex\hbox{E}\kern-.125emX}}

\let\svthefootnote\thefootnote

\begin{document}

\title{GenSyn: A Multi-stage Framework for Generating Synthetic Microdata using Macro Data Sources\\

\thanks{This research was supported in part by the NSF grants DGE: 1922598 (PI: Sikdar) and 1945764 (PI: Sikdar).}
}

\author{Anonymous authors}
\author{\IEEEauthorblockN{Angeela Acharya}
\IEEEauthorblockA{\textit{George Mason University} \\
Fairfax, Virginia, USA \\
aachary@gmu.edu}
\and
\IEEEauthorblockN{Siddhartha Sikdar}
\IEEEauthorblockA{\textit{George Mason University} \\
Fairfax, Virginia, USA \\
ssikdar@gmu.edu}
\and
\IEEEauthorblockN{Sanmay Das}
\IEEEauthorblockA{\textit{George Mason University} \\
Fairfax, Virginia, USA \\
sanmay@gmu.edu}
\and
\IEEEauthorblockN{Huzefa Rangwala}
\IEEEauthorblockA{\textit{George Mason University} \\
Fairfax, Virginia, USA \\
rangwala@gmu.edu}}


\maketitle

\begin{abstract}
Individual-level data (microdata) that characterizes a population, is essential for studying many real-world problems. However, acquiring such data is not straightforward due to cost and privacy constraints, and access is often limited to aggregated data (macro data) sources. 
In this study, we examine synthetic data generation as a tool to extrapolate difficult-to-obtain high-resolution data by combining information from multiple easier-to-obtain lower-resolution data sources. In particular, we introduce a framework that uses a combination of univariate and multivariate frequency tables from a given target geographical location in combination with frequency tables from other auxiliary locations to generate synthetic microdata for individuals in the target location. Our method combines the estimation of a dependency graph and conditional probabilities from the target location with the use of a Gaussian copula to leverage the available information from the auxiliary locations. We perform extensive testing on two real-world datasets and demonstrate that our approach outperforms prior approaches in preserving the overall dependency structure of the data while also satisfying the constraints defined on the different variables.
\end{abstract}


\section{Introduction}
Access to comprehensive microdata that correctly captures the important characteristics of a population is essential in addressing critical societal needs.
For instance, in epidemiology, it is needed for identifying disease progression, modeling interactions, or testing interventions \cite{simulated_maxentropy, Lee2010-vj}. 
However, acquiring such data is not always straightforward due to constraints on cost, policy, and data sharing.
Additionally, merging disparate data sources to get all the required information in a single dataset is hindered by the difficulty of obtaining linked data with shared attributes. Often, key information is only available at an aggregated level in terms of summary statistics. 
One solution to the problem of acquiring representative microdata is to turn to \emph{synthetic data}, which is artificially generated rather than captured from real-world events. 

Synthetic population generation (or population synthesis) is the process of combining multiple datasets from different sources and granularities to obtain individual-level data representing population attributes \cite{li2020sync}. It is a fundamental step in many applications such as microsimulation \cite{microsimulation}, which involves simulating large-scale social systems such as disease outbreaks and transportation planning.
In these applications, the datasets to be generated can often be represented with categorical data attributes \cite{simulated_maxentropy}. 

Prior work on population synthesis typically relies on the availability of sample individual-level data. This allows for the extraction of necessary relationships such as correlation structure and marginal distributions.
Properties of the complete population data are then extrapolated using macro data as well as this information extracted from properties of the sample data. Synthetic Reconstruction (SR) \cite{SR} and Combinatorial Optimization (CO) \cite{CO} are two of the most widely used techniques for population synthesis.
On the other hand, recent work on synthetic data generation involving techniques such as Generative Adversarial Networks (GANs) \cite{NIPS2014_5423} and Variational Auto-Encoders (VAE) \cite{8285168} is typically applicable in situations where the data is available but cannot be publicly released, or is insufficient in scale. 
Our focus in this study is on situations where individual-level real data is not available at all and microdata with categorical attributes from multiple sources need to be generated solely based on macro data sources.
Using macro data alone for population synthesis is a challenging problem and only a few prior studies have tried to tackle it, with limited success \cite{taxman2013simulation, synthacs, li2020sync}.
\let\thefootnote\relax\footnote{978-1-6654-8045-1/22/\$31.00 ©2022 IEEE}

Our central hypothesis is that it is possible to leverage information contained in macro datasets (univariate/multivariate frequency distributions) for estimating the desired joint probability distribution of features over individuals. We evaluate this estimation by checking if it preserves the overall dependency structure of the data while also satisfying the constraints defined on the variables.

We propose a novel multi-stage framework, GenSyn, which combines a Gaussian copula approach with a method for generating 
dependency graphs based on conditional probabilities estimated from univariate and multivariate frequency distributions. The main objective is to generate synthetic population data for any given geographical location. 
These methods independently handle different aspects of the synthetic data generation problem. The estimation of a dependency graph with appropriate conditional probabilities models the multivariate relationships specific to the target population, as specified in the macro data. However, since the overall joint distribution information is usually not available from the macro data belonging to a single geographical location, this method can only capture partial dependencies (see Section \ref{conditional}). The copula sampling method (see Section \ref{copula}) helps to overcome this by providing an estimation of the joint distribution using the dependency information captured from multiple geographical locations. Both these methods independently estimate the probability distribution of the desired population which is later averaged and further optimized using maximum entropy optimization to jointly satisfy location-specific marginal constraints. Together, these methods ensure that the generated data satisfies the following criteria: i) maintaining the overall dependency structure of the data, ii) satisfying the bi-variate conditional relationships of the variables, iii) satisfying the uni-variate marginal constraints.
\addtocounter{footnote}{-1}\let\thefootnote\svthefootnote

The main contributions of this study are:
\begin{itemize}
\item We propose a novel multi-stage framework that combines categorical attributes from multiple macro data sources and generates synthetic population data with attributes from all these sources without requiring any real microdata.
\item We perform extensive validation experiments and demonstrate that our approach generates quality synthetic data that outperforms prior approaches when considering a variety of metrics on two real-world datasets.
\item We make all the code and data for GenSyn publicly available for others to use in similar tasks, and to encourage reproducibility and transparency.\footnote{https://github.com/Angeela03/GenSyn}
\end{itemize}

\begin{figure*}[t]
\centering
  \includegraphics[width=0.6\linewidth]{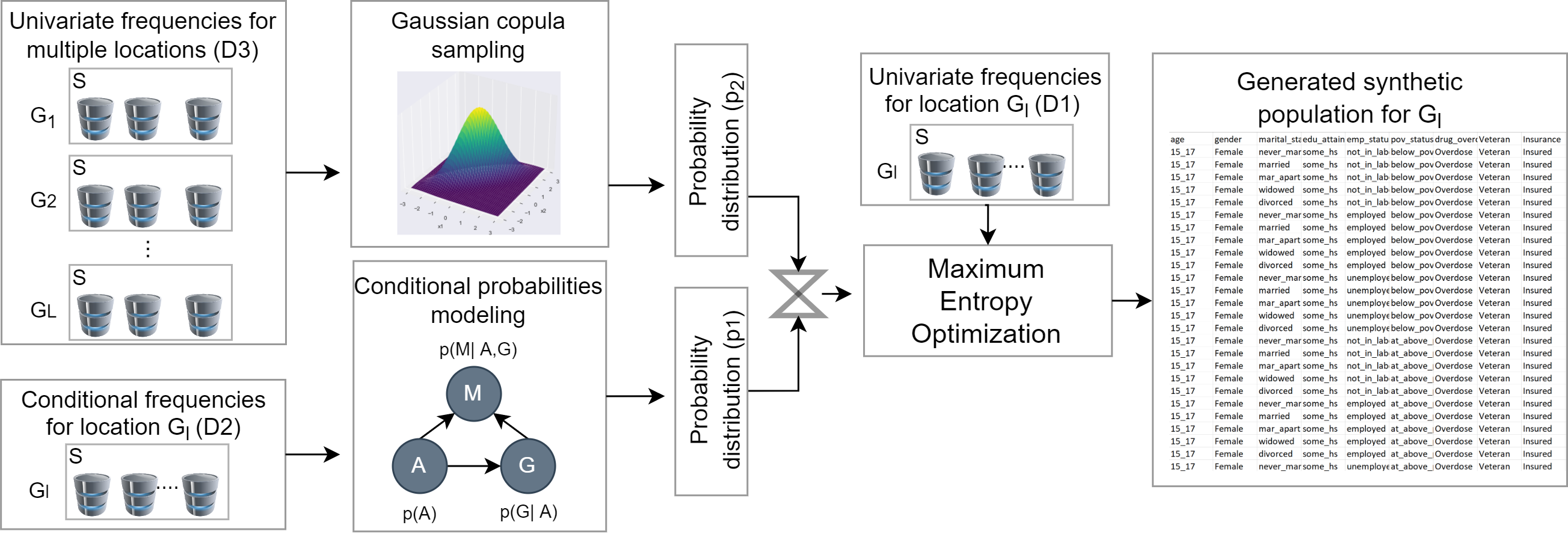}
\caption{High-level overview of the proposed approach}
\label{fig:method}
\end{figure*}

\section{RELATED WORKS}

\subsection{Synthetic Reconstruction}
Synthetic Reconstruction (SR) is the most common category of techniques for population synthesis. The standard approach for building synthetic populations is based on a technique called Iterative Proportional Fitting (IPF) developed by Beckman et al. \cite{BECKMAN1996415} and involves random sampling from a series of conditional probabilities derived from contingency tabulations capturing joint probability distributions. Prior research has used this approach to generate synthetic populations for studying large-scale societal phenomena such as micro-simulation for transportation study and land use \cite{article, wheaton_synthesized_2009}. Several variations of IPF have been proposed over the years \cite{IPF-variant1, IPF-variant2}. IPF-based approaches have many advantages but they usually require a seeding population (sample data) as input which is not always available \cite{doi:10.1177/0361198120964734}. 

Some SR-based methods like maximum entropy optimization \cite{Lee2011CrossEntropyOM} can generate a synthetic population solely based on the constraints defined on a population and do not always require a seeding population. However, the dependency structure of the data is usually not preserved in this case which is a serious limitation. SR-based methods are deterministic and the results do not vary significantly across each run.

\subsection{Combinatorial Optimization}
Unlike SR, Combinatorial Optimization (CO) methods iteratively select the best combination of individuals for a population from a subset of the population by minimizing a fitness function and are stochastic in nature \cite{CO}. This includes techniques like Simulated Annealing \cite{harland2012}, Hill climbing \cite{Abraham2012PopulationSU}, and Genetic Algorithm \cite{Birkin06asynthetic}. These approaches also normally require sample data as input. Compared to SR, CO methods usually provide a more accurate estimate of the population. However, as the population size grows, these methods cannot guarantee an optimal solution and the computational complexity grows exponentially \cite{doi:10.1177/0361198120964734}.
\subsection{Copula}
Recently, copula-based methods have gained popularity in the synthetic data generation community due to their ability to model dependencies between random variables. For instance, Benali et al. \cite{benali:hal-03188317} proposed MT-Copula that generates complex data with automatic optimization of copula configurations. Most of the prior approaches based on Copula are restrictive for our application because they rely on the availability of real data to model the dependency structure of the required synthetic data. Zheng et al. \cite{li2020sync} proposed an alternative approach based on Gaussian Copula called SynC, which is less restrictive and can generate microdata based on macro data sources. However, they noted that their approach does not produce good reconstruction accuracies on large population sizes. Thus, it is not directly applicable to this problem and has to be combined with other methods to produce a better performance (as verified in Section \ref{results_and_discussion}).

\subsection{Machine Learning}
Different supervised and unsupervised machine learning techniques have been implemented for synthetic data generation. For instance, Choi et al. \cite{DBLP:journals/corr/ChoiBMDSS17} proposed a method called MedGAN to generate longitudinal EHR data for patients. SynthPop \cite{synthpop} implemented techniques like SVM, logistic regression, and CART to generate synthetic data for preserving disclosure risk. Although these studies have demonstrated capabilities of producing high-quality synthetic data, they normally rely on the availability of some form of real microdata for training which is beyond the focus of this study.

\subsection{Hybrid Methods}
Recent research has suggested that combining different methods proves to be the most relevant means for generating synthetic population in some configurations \cite{doi:10.1177/0361198120964734}. For instance, SynthACS \cite{synthacs} borrowed ideas from conditional probabilities \cite{doi:10.1068/a201645}, deterministic re-weighting \cite{doi:10.1002/psp.351}, and simulated annealing \cite{harland2012} to generate synthetic population data using American Community Survey (ACS) dataset. While the combined approach performs better than the participating individual approaches and many prior approaches, one major drawback of SynthACS is its inability to properly maintain the dependency structure of the data (see section \ref{results_and_discussion}). Nevertheless, the SynthACS framework, which is based on R, provides scalable tools for researchers to generate synthetic population data for any geographical location in the United States. We make use of this framework and extend it for our application.
Our approach, which also combines multiple methods, resolves many of the prior limitations by allowing us to incorporate macro data from multiple geographical resolutions to generate synthetic data for a target geographical location.

\section{METHODOLOGY}
\subsection{Problem Formulation}
We assume that the input macro data comes from multiple sources. Let S denote the set of all sources. The goal of GenSyn is to combine information from all sources in S based on the input macro data to construct synthetic microdata with attributes from all sources. Let X = $\{X_1, X_2,...X_K\}$ represent the set of all possible attributes (variables) in S with K being the total number of attributes. For simplicity, we consider the case of generating synthetic microdata for a single geographical location $G_l$. The assumption is that the univariate and bivariate (or multivariate in some cases) distributions of the population attributes are available to us. We distinguish between the different types of available macro data as follows (more details in Section \ref{section-macrodata}):
\begin{itemize}
\item D1: Univariate frequency distributions of variables in S for the target location $G_l$. E.g. count of the total number of males and females present in Baltimore County, MD.
\item D2: Bivariate or multivariate frequency distributions of variables for the target location $G_l$. E.g. age by gender counts for the population in Baltimore County, MD.
\item D3: Univariate frequency distributions of variables in S for multiple auxiliary geographical locations $G_0, G_1,....G_L$. E.g. number of people living in poverty, number of people with high school degrees across different counties in the United States. This is similar to D1 but is for multiple locations.
\end{itemize}

 Let $c_k = \{c_k^{1}, c_k^{2},..., c_k^{R}\}$ represent the set of all possible categories of a random variable $X_k$, where R is the total number of categories. We represent a vector of categorical random variables by a tuple T = $(x_1,x_2,...x_K)$ where $x_k$ $\epsilon$ $c_k$ represents the value of an attribute $X_k$. The space of all possible categorical tuples is represented by $\mathbf{S_c}$ = $\varprod_{k=1}^K c_k$, where $\varprod$ represents the cartesian product over $c_k$. Microdata for a location $G_l$ is denoted by an N $\times$ K dimensional matrix $M = ((x_{1}^{1},x_2^{1},...x_K^{1}), (x_{1}^{2},x_2^{2},...x_K^{2}) ,...
 (x_{1}^{N},x_2^{N},...x_K^{N}))$, where a tuple $(x_{1}^{n},x_2^{n},...x_K^{n})$ represents the characteristics of an individual ``n" in the population (population profile) and N denotes the size of the population.

\subsection{Architecture}
The overall architecture of GenSyn is depicted in Fig. \ref{fig:method}. As shown, GenSyn consists of three major components: i) Conditional probabilities modeling, ii) Gaussian copula sampling and iii) Maximum entropy optimization, which is discussed in detail in the following subsections. 
On a high level, this framework first uses Conditional probabilities modeling and Gaussian copula sampling in parallel to estimate the joint distribution of the tuple space $S_c$ as defined by macro data D2 and D3 respectively. 
Both these methods independently return probability distributions (p1 and p2) for $S_c$. We then take the average of p1 and p2 to get a single probability distribution $p(X_1,..X_{i-1}, X_K)$ for the profiles in $S_c$. Finally, we use maximum entropy optimization to further refine this probability distribution based on data D1. Let $W = \{w_1, w_2,...w_N\}$ represent the final probability distribution returned after the optimization. We can now obtain the prevalence (frequency of occurrence) of the N profiles in $S_c$ by multiplying the probability values in W by the size of the population N. Replicating the profiles based on their prevalence values yields the desired synthetic population data M. 

\subsubsection{Conditional Probabilities Modeling}\label{conditional}
A good quality synthetic data needs to maintain the conditional relationships between different variables that exist in the macro data.
As mentioned above, D2 consists of bivariate or multivariate conditional frequencies of variables for a target location $G_l$. To best represent this information, we first generate a directed acyclic dependency graph (DAG) $\mathcal{G}$ for the variables using the conditioning relationships that are available from the macro data sources (see Section \ref{section-macrodata}). Each node in the graph represents a variable and the edges represent variable conditioning. A node ``a" that has a directed edge to some other node ``b" is referred to as the parent node of ``b". A dependency graph constructed based on the above example is shown in Fig. \ref{fig:method}. 

We use $\mathcal{G}$ to estimate the joint probability distribution (p1) of $S_c$ using ideas from the conditional probabilities method. First introduced by Birkin and Clarke \cite{doi:10.1068/a201645}, this method computes multivariate joint distribution using probabilities implied by the macro data. As outlined in Algorithm 1, we first define an ordering on the variables based on $\mathcal{G}$. Then, we proceed by estimating the joint distribution of the categorical space returned by the first two variables and iteratively adding new variables conditioned based on one or more of the previously added ones until the joint distribution p1 is returned. The order in which the variables are added is important because variables in the parent nodes should always come before any variable in the child node. This algorithm also allows adding variables that are independent of the rest of the variables in the data. For an independent variable $X_i$, we define the joint probability $p(X_1,..X_{i-1}, X_i)$ as $p(X_i) \times p(X_1,..,X_{i-1})$. The creation of dependency graphs and ordering of variables is explained in more detail in the Appendix.

\SetKwComment{Comment}{/* }{ */}
\begin{algorithm}[t]
\caption{Modeling conditional relationships}
\label{alg1}
\KwIn{Macro data D2, Dependency graph $\mathcal{G}$}
\KwOut{Joint probability distribution $p_2$ for $S_c$}
K $\gets$ Number of variables\\
Order the variables based on $\mathcal{G}$. \\
For the first two variables, X1 and X2, calculate the joint probability distribution $p(X_1,X_2)$ based on the macro data\\
i $\gets$ 3\\
\While {i $\leq$ K}
    {$p(X_1,..X_{i-1}, X_i) \gets p(X_i | p_{xi}) *p(X_1,..,X_{i-1})$ \Comment*[r]  {where $p_{xi}$ $\subseteq$ $\{X_1,..,X_{i-1}\}$ are the parent nodes of $X_i$ in $\mathcal{G}$ }
    }
\Return {$p_1 = p(X_1,..X_{i-1}, X_K)$} 
\end{algorithm}

\subsubsection{Copula Sampling}\label{copula}
Ideally, a DAG representing the joint distribution of the real individual-level data could be used for estimating the synthetic population distribution. When that is not available, information has to be extracted from the available macro data sources. Publicly available macro data for a geographic location $G_l$ usually only consists of univariate frequency distributions and some multivariate conditional relationships between variables which we modeled using the conditional probabilities method. However, that alone is not sufficient to satisfy the overall dependency structure of the data (see section \ref{association_plots}). To overcome this, we make use of information from multiple geographical locations $G_0, G_1...G_L$ (D3) and model the dependency structure of the variables in $S_c$ using a Gaussian Copula to estimate the probability distribution of the population at $G_l$.

The term copula was first employed (in a mathematical sense) by Sklar \cite{sklar1959fonctions} to describe functions that join (or couple) multiple one-dimensional distribution functions to form a multivariate distribution function \cite{10.5555/1204326}. According to Sklar's theorem, any multivariate joint distribution can be written in terms of univariate marginal distribution functions and a copula that describes the dependency structure between the variables. This theorem also states that there exists a unique copula for any set of random variables with continuous cumulative distribution functions (CDFs). Let F $\epsilon$ $F(F_1, . . . , F_M)$  be an M-dimensional distribution function with marginal distribution functions $F_1,...,F_M$. Then, there exists a copula C for a random vector $(X_1,...,X_M)$ such that: $F(X_1,...,X_M) = C(F_1(X_1),...,F_M(X_M))$

Let, $(U_1,...,U_M) = (F_1(X_1),...,F_M(X_M))$. Then, the copula of $(X_1,...,X_M)$ is defined as the joint cumulative distribution function:
$C(u_1,...,u_M) =  Pr[U_{1}\leq u_{1},...,U_{M}\leq u_{M}]$

Motivated by SynC \cite{li2020sync}, we implement a Gaussian Copula to model the probability distribution of the variables based on the covariance matrix $\Sigma$ obtained from D3 and the individual marginals of the components in D3. The Gaussian copula is the most prominent type for copula \cite{ROSS201397} because of its simplicity and interpretability and has been widely used to model multivariate dependency structure. For a given $\Sigma$, a Gaussian Copula can be written as: \[C(u_1,...,u_M) = \phi_{\Sigma} (\phi^{-1}(u_1),...,\phi^{-1}(u_M))\]
where $\phi_{\Sigma}$ represents the CDF of a multivariate 
normal distribution with zero mean, unit variance, and correlation coefficient $\Sigma$ and $\phi^{-1}$ represents the inverse CDF of the marginal distributions. We
assume M to be the count of the total number of variable categories
(i.e. M = $\sum_{k=1}^K |c_k|)$. D3 consists of frequency counts of every categorical component represented by M for locations $G_0,G_1,...G_L$. We can think of it as a matrix with rows representing the different locations and columns representing the categorical components. We first convert the frequency counts to percentages before passing them to the copula sampling algorithm because we want the data to be normalized for modeling the distributions. We use a beta distribution (with parameters $\alpha$ and $\beta$) to model the marginal distributions for the categorical components such that: $
\frac{\alpha}{\alpha + \beta} = \mu_{l}^{m}$ and $\frac{\alpha\beta}{(\alpha + \beta)^2 (\alpha+\beta+1)} = (\sigma_{l}^{m})^2$. 
Here, the mean ($\mu_{l}^{m}$) and variance ($(\sigma_{l}^{m})^2$) of component m for a location l can be estimated using macro data D3 by computing the mean and variance of the component across multiple locations. It is important to note that even though the variables are categorical, the frequency values represented by the individual categorical components are continuous. Thus, the beta distribution assumption is valid. The process of obtaining a joint distribution (p2) using a Gaussian Copula is provided in Algorithm 2.


\begin{algorithm}[h]
\caption{Gaussian Copula Sampling}
\label{Algo2}
\KwIn{Macro data D3}
\KwOut{Joint probability distribution (p2) for $G_l$}
M $\gets$ Count of the total number of variable categories
(i.e. M = $\sum_{k=1}^K |c_k|)$ \\
D $\gets$ Normalized macro data D3 of dimension $L \times M$, where L is the total number of auxiliary geographical locations\\
it $\gets$ Number of iterations \Comment*[r] {Samples are generated ``it" times and the average probability distribution is returned to get an accurate estimation of the distribution}
$\Sigma$ $\gets$ M $\times$ M covariance matrix of D\\
$\phi$ $\gets$ CDF of a standard normal distribution\\
$F_m^{-1}$ $\gets$ Inverse CDF of the marginal distribution (beta distribution) of the $m^{th}$ component of D\\
list\_prob $\gets$ []\\
\For{i = 1 to it} 
{
Draw $Z = Z_1, Z_2,.....Z_M$ $\epsilon$  N(0, $\Sigma$), an M-dimensional normal distribution with mean 0 \& covariance matrix $\Sigma$ \\
\For{m = 1 to M}{
 $u_m$ $\gets$ $\phi(Z_m$)\\
 $y_m$ $\gets$ $F_{m}^{-1}(u_m)$
 }
 Y $\gets$ $\{y_j\}_{j=1}^M$\\
 Sample from Y using Algorithm 3 (see Appendix) to get a joint probability distribution $p_i$ = $p(X_1,..X_{i-1}, X_K)$\\
 list\_prob.append($p_i$)
}
Average the probability distributions in list\_prob to get $p_2 = p(X_1,..X_{i-1}, X_K)$\\
\Return {$p_2$}
\end{algorithm}

\subsubsection{Maximum Entropy Optimization}\label{entropy}
Algorithms 1 and 2 independently return probability distributions p1 and p2 of the population profiles in $G_l$ based on population-specific data (D2) and more generic data (D3), respectively. We combine the probability values returned by these methods to obtain a new distribution p for the profiles in $S_c$ by taking an average of p1 and p2. This is to ensure that the desired synthetic data satisfies the necessary dependencies. However, at this point, the joint distribution p does not satisfy the required population marginal constraints. Thus, we use the principle of maximum entropy to return the final probability distribution of the population profiles (or the corresponding weights W) satisfying the given constraints of the population at $G_l$.

\textit{Defining constraints.}
Any piece of information from the macro data that our synthetic data must satisfy can be incorporated as a constraint. For instance, a constraint for satisfying the frequency counts of an attribute $X_k$ can be defined as: $\sum_{i=1}^{N} w_i \mathds{1}$ ($x_k^{i} == x) = \eta$, where the left-hand term denotes the percentage of profiles in $S_c$ that has a value x for a variable $X_k$ and the right-hand term ($\eta$) denotes the evidence that we want our microdata to be consistent with i.e. the actual percentage of profiles with that value. 
$\mathds{1} (x_{k}^{i} == x)$ is an indicator function that indicates whether the value of a variable $X_k$ is equal to some value x for the $i^{th}$ population profile. 
\[ \mathds{1} (x_{k}^{i} ==x)  = \begin{cases} 
    1 & \text{if $x_{k}^{i}$ = x} \\
    0 & \text{otherwise}
    \end{cases}\]
    
We define J such constraints that we want to satisfy which can be written in a more generic form as: 
$\sum_{i=1}^{N} w_i f_j(T_i) = \eta_j,  \forall j \in J $ where $f_j(T_i)$ denotes an indicator function relating to a constraint j defined over a tuple $T_i$.\\

\textit{Principle of maximum entropy.}
In practice, many possible joint probability distributions can fit a given set of constraints. The principle of maximum entropy (minimum cross entropy) states that the distribution with the largest entropy is the one that best represents the available information, with respect to some prior knowledge \cite{e3030191}. In this context, the weights (W) should be as uniform as possible or should be close to the priors while being consistent with the defined constraints \cite{taxman2013simulation}.

This is equivalent to minimizing:
   \[ L = \sum_{i=1}^{N}[ w_i \log \frac{w_i}{u_i}]  + \sum_{j}^{J} \theta_j [\eta - \sum_{i=1}^{N} w_i f_j(T_i)] \]
    where the first term represents the divergence of the estimated probabilities from a prior set of probabilities U and the second term represents the constraints underlying the data. Here, $\theta_j$ is a Lagrange multiplier. The joint probability distribution (p) obtained from the previous step is used as the prior (U) in this equation.
    We can solve this primal Lagrange function by setting $dL/ dw_i$ = 0. This gives: 
    $ w_i = u_i \exp(\sum_{j}^{J}f_j(T_i)\theta_j - 1 )$. Plugging the solution back into the primal function, we get a dual function $L = \sum_{j}^{J}\theta_j\eta_j - \sum_{i=1}^{N} u_i \exp(\sum_{j}^{J}f_j(T_i)\theta_j -1 )$.
   Now, we maximize this function and get the optimal values for the Lagrange multipliers $\theta$ which is equivalent to minimizing the negative of the function:
   \[ \operatorname*{argmin}_\theta [
 -\sum_{j}^{J}\theta_j\eta_j + \sum_{i=1}^{N} u_i \exp(\sum_{j}^{J}f_j(T_i)\theta_j - 1) ]\]
 

   
 
Finding the optimal $\theta$ and plugging the solution to the equation of $w_i$ can return the optimal values of W which in turn can be used to estimate the population of $G_l$. The optimization was performed using the L-BFGS-B algorithm \cite{Zhu94l-bfgs-b}.

In practice, modeling multi-dimensional categorical data using maximum entropy can be infeasible due to exponential possibilities of attribute tuple space \cite{simulated_maxentropy}. We resolve this by limiting the number of categorical spaces to be explored using the results from probability distribution p. To be specific, we omit the tuples that have a probability less than some value $\tau$. $\tau$ is a parameter of the model whose value is decided based on experiments (see Section \ref{tauexperiments}).

\section{Experiments}
\subsection{Data Description}

\subsubsection{Macro Data}\label{section-macrodata}
We experimented with two different sets of macro data for this study. 

\textbf{Demographic data (ACS).} 
Macro data comprising important population demographics were collected from the American Community Survey (ACS) \cite{noauthor_acs_nodate}. ACS has been widely used in synthetic data generation studies \cite{simulated_maxentropy, synthacs}.
To be able to compare our performance with a prior method (SynthACS) based on ACS tables, as well as based on what is available in the validation dataset, we selected the following variables from ACS: age, gender, marital status, education attainment, employment status, poverty status, nativity status, and geographic mobility. To ensure that our approach generalizes well, synthetic microdata with demographic attributes were generated for 50 different counties across the United States. 

\textbf{Opioid overdose-related data (Combined).} 
In addition to the demographic data, a domain-specific dataset was also formulated to examine GenSyn's ability to combine macro data across multiple sources. One particular application that motivated our study was the opioid overdose epidemic. Opioid overdose is a complex health problem that presents a significant burden across people of all demographic and socioeconomic categories. 
Addressing this problem requires the study of data across multiple domains (i.e., socio-economic, demographic, health-related) \cite{Jalali2020}. We collected macro data from these domains to formulate our dataset. This includes sources like ACS \cite{noauthor_acs_nodate}, CDC Wide-ranging Online Data for Epidemiologic Research (CDC WONDER) \cite{cdc_wonder}, Small Area Health Insurance Estimates (SAHIE) \cite{bureau_2021}, and Opioid Environment Policy Scan (OEPS) \cite{oeps}. The variables included in this dataset are age, gender, marital status, education attainment, employment status, poverty status, insurance, opioid overdose, and veteran population. Synthetic microdata for these variables was generated for 50 counties in the United States with the highest overdose counts.

Table \ref{tab:data} provides a description of the variables included in these datasets. Here,``Categories" represents the number of variable categories, and ``Conditioning" represents the variable conditioning information that is obtained from the macro data. For instance, gender is said to be conditioned on age if gender by age counts are accessible from the macro data \cite{synthacs}. 
\setlength{\tabcolsep}{0.2pt}
\begin{table}[h]
\caption{Description of variables used in our datasets}
\begin{center}
\begin{tabular}{cccc}
 Attribute & Source & Conditioning & Categories\\
 \hline
 Age & ACS & Gender & 16\\
 Gender & ACS & Age & 2\\
 Marital status & ACS & Age \& Gender & 5\\
 Education attainment & ACS & Age \& Gender & 7\\
 Employment status & ACS & Age \& Gender & 3\\
 Poverty status & ACS & Gender \& Emp status & 2\\
 Nativity status & ACS & Age & 2\\
 Geographic mobility & ACS & Edu attain & 3\\
 Insurance & SAHIE & Gender & 2\\
 Opioid overdose & CDC Wonder & Age & 2\\
 Veteran population & OEPS & Gender & 2\\
 \hline
\end{tabular}
\end{center}
\label{tab:data}
\end{table}
\subsubsection{Microdata for Validation}
We generate synthetic microdata using macro data sources with the assumption that any kind of microdata is not accessible. While population-level microdata comprising of attributes from multiple sources is not publicly available, a sample of the population data is made available by ACS every year as Public Use Microdata Sample (PUMS) \cite{noauthor_acs_nodate}. This sample data consists of individual-level de-identified records for the ACS variables and is a good representation of the actual census data. For each county under study, we use the corresponding individual-level sample from PUMS to assess the performance of our approach and the baselines.  Since PUMS only consists of demographic data, we use it only for the ACS variables. 

\subsection{Data Processing}
The attributes used in this study come from multiple sources. Thus, they differ in their formats. However, since each attribute is conditioned based on some other attributes, their formats should be consistent across multiple sources. For instance, ACS uses 5-year age groups for its age attribute. Thus, if any attribute is conditioned based on age (e.g., Opioid overdose), the age categories have to be the same as in the ACS data. However, since that is not always the case, extensive data preprocessing was done to ensure that all the attributes followed a similar format. Similarly, to preserve the meaningfulness of the generated synthetic data, a few assumptions had to be made about the attributes to be merged. For instance, all individuals under 15 were assumed to have never been married. The data formatting steps and the assumptions on the attributes are similar to the ones specified in \cite{synthacs}.

\subsection{Baselines}
While there are many approaches to synthetic data generation, only a few apply to the problem of generating population data from aggregated sources. Our baselines focus on those.

\begin{itemize}
    \item Max entropy: This is the maximum entropy optimization framework without any prior \cite{Lee2011CrossEntropyOM}. This method returns synthetic data that best fits the given marginal constraints without making any assumption about the data distribution.
      $ L = \sum_{i=1}^{N}[ w_i \log w_i]  + \sum_{j}^{J} \theta_j [\eta - \sum_{i=1}^{N} w_i f_j(T_i)] $
    \item Conditional: In this case, we generate a synthetic population using just the conditional probabilities method \cite{doi:10.1068/a201645}. Essentially, once the joint distributions are returned by Algorithm 1, we sample from that distribution to generate synthetic population data.
    \item SynC: This is the implementation of the Copula Sampling method from the paper SynC \cite{li2020sync}. However, one difference with this approach is that since we do not have a large number of attributes, we sample all attributes at once rather than in batches.
    \item SynthACS: This is the implementation of the paper SynthACS \cite{synthacs}. 
    \item Syntropy: This is the baseline we created by combining the conditional probabilities approach and the maximum entropy approach. Basically, instead of combining both Copula and Conditional probabilities to generate the joint distributions, the maximum entropy optimization only uses the probabilities returned from the conditional probabilities algorithm as its prior.
\end{itemize}

  \begin{figure*}[t]
\centering
 \includegraphics[width=1\linewidth]{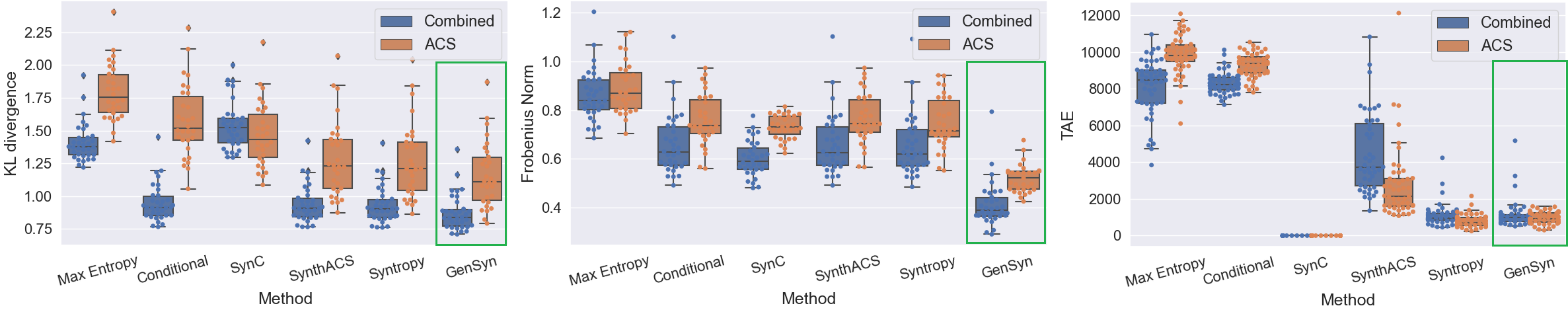}
\caption{Comparison of GenSyn (highlighted in green) with the baseline approaches based on KL divergence, Frobenius norm, and TAE: Each data point represents a county and the colors differentiate the two datasets (ACS and Combined)}
\label{fig:measures}
\end{figure*}

\subsection{Evaluation}
We evaluate our method using the following metrics:
\subsubsection{Aggregated Statistics}
The aggregated statistics of the variables in the synthetic data should be close to the statistics of the variables in the macro data. This is measured in terms of Total absolute error (TAE) which is defined as:
$ TAE = \sum_{k}^{K}\sum_{j}^{|c_k|}|O_{jk} - E_{jk}|$,
where $O_{jk}$ and $E_{jk}$ are the observed (generated) and expected (actual) counts of categorical frequencies for the attributes X = $\{X_1,X_2,...X_K\}$, respectively. Here, $|c_k|$ denotes the total number of categories in an attribute $X_k$. We calculate the TAE values for the 50 counties for both datasets. 
A lower TAE value signifies that the synthetic data is better at preserving the marginal constraints.

\subsubsection{Probability Distribution}
The overall distribution of the generated synthetic data should be similar to the expected distribution. This is measured by Kullback-Leibler (KL) divergence (or relative entropy) \cite{Joyce2011}. Formally,
$D_{KL} (P||Q) =  \sum_{x \epsilon X} P(x) \log (\frac{P(x)}{Q(x)})$, where P represents the reference distribution and Q represents the distribution of the synthetic data. The reference probability distribution is calculated using the PUMS data.
A lower KL divergence value signifies that the overall distribution of the generated synthetic data is closer to the expected distribution. 

\subsubsection{Pairwise Relationships}
The association (or correlation) between variables in the synthetic data needs to be consistent with the association in the real data: measured using Cramer’s V \cite{inbook} for categorical variables. We generate association matrices for the different approaches and compute their similarity with the association matrix of the sample population using the Frobenius norm: d(A,B) = $\sqrt{\sum_{i=1}^{K}\sum_{j=1}^{K} (a_{ij} - b_{ij})^2}$, where A is the expected association matrix, B is the observed association matrix and K is the number of variables. The lower Frobenius norm value signifies that the dependency structure of the synthetic data is similar to the expected dependency structure.

\subsection{Experiments with $\tau$}\label{tauexperiments}
As mentioned in Section \ref{entropy}, the parameter $\tau$ is used to limit the number of categorical spaces to be explored by the maximum entropy model. We only select the tuples that have a probability greater than or equal to $\tau$. We later normalize the probability values for them to be in the range of 0 to 1.

We experiment with different values of $\tau$ and observe how the overall KL divergence changes. We start with $\tau = \frac{1}{10^{-1} \times N}$ (where N is the size of the population) and keep dividing the value by 10 until we reach the minimum probability value, to increase the number of categorical tuples to be explored. We also visualize how the $\tau$ values affect the performance of infrequent categories in the synthetic data. For this, we choose the categories that appear the least number of times in both datasets: \% of people with opioid overdose (Combined), and \% of people who moved from abroad (ACS).

\section{Results and Discussions}\label{results_and_discussion}
\subsection{Performance Evaluation and Comparison}
We assess the performance of the proposed approach (GenSyn) and the baselines based on KL divergence, Frobenius norm of the association matrices, and TAE (see Fig. \ref{fig:measures}). As shown in the figure, our approach performs better than the baselines in terms of KL divergence and Frobenius norm for both datasets (ACS and Combined). This indicates that the distribution of data generated by GenSyn is the closest to the expected distribution and that GenSyn better preserves the dependencies between the variables. In terms of TAE values, SynC and Syntropy were found to perform better on average. In fact, SynC returned 0 TAE values for all counties because the algorithm for SynC exactly matches the marginal distributions of the variables. However, while doing so, many of the variable dependencies are lost, resulting in higher KL divergence and Frobenius norm values as shown in the figure. For the synthetic data generation process, it is important to evaluate the quality of data using all the three measures together. Therefore, based on the results, GenSyn is the best choice for correctly satisfying the dependency structure of the data while also maintaining low marginal errors.


\subsection{Association Plots}\label{association_plots}
\begin{figure}[t]
\centering
 \includegraphics[width=0.95\linewidth]{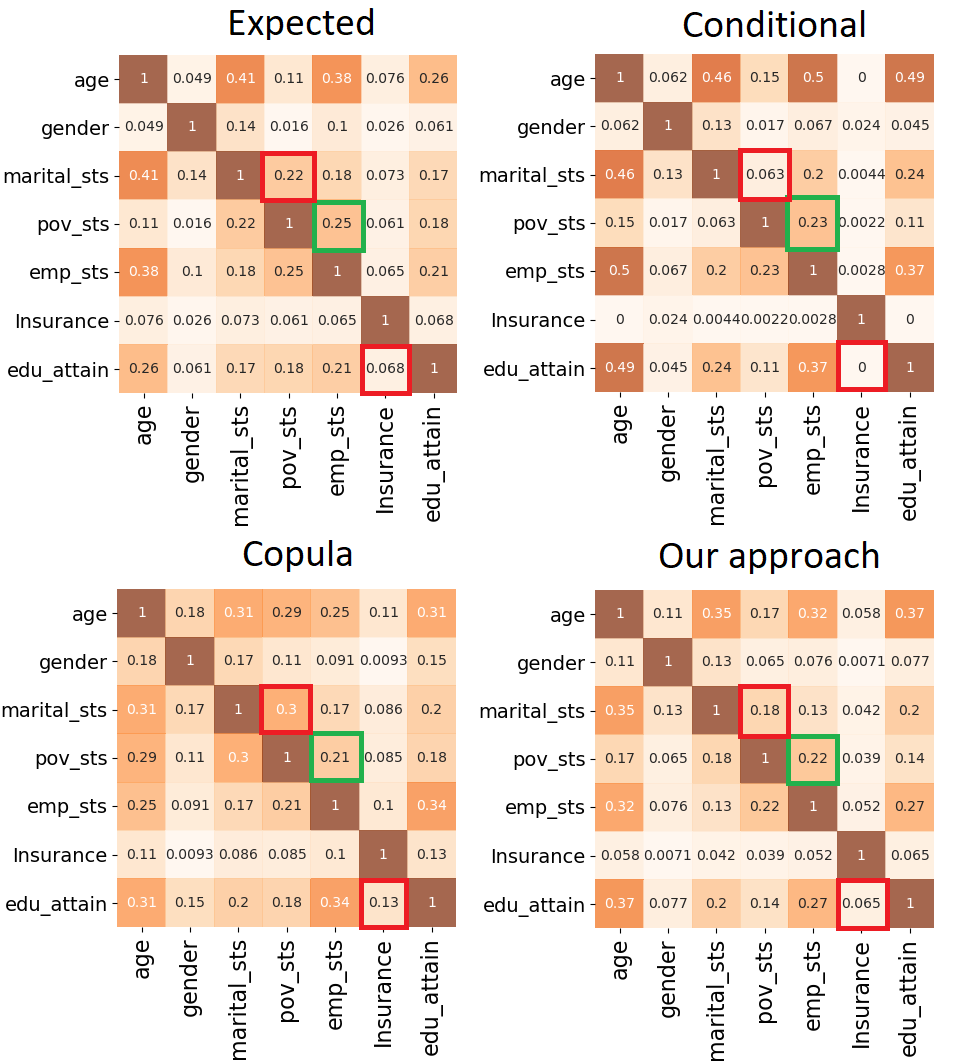}
\caption{Example of association matrices obtained using different methods for a single county}
\label{fig:pop_corr}
\end{figure}

We noted from Fig. \ref{fig:measures} that the Frobenius norm values returned by GenSyn are much lower than the baselines meaning that the dependency relationships are better preserved by GenSyn. We plot the association matrices returned by GenSyn along with the ones returned by the two approaches (Conditional probabilities and Copula) used to formulate it and visualize how they differ from the expected associations. We present the results from a single county as an example. As illustrated in Fig. \ref{fig:pop_corr}, the association values returned by the conditional probabilities method are close to the expected values for some variable pairs, especially the ones whose bivariate frequencies are available in the macro data (e.g. (pov\_status, emp\_status) -- highlighted in green). However, one of the major limitations of the conditional probabilities method and the prior approaches that are built upon it (e.g. SynthACS) is their inability to correctly infer the dependencies for the variable pairs whose bivariate frequencies are not available in the macro data. As an example, we refer to a variable pair that is highlighted in red in the figure (pov\_status, marital\_status). This variable pair has a high association value in the expected data (0.25). But the conditional probabilities method could not correctly infer that association because pov\_status by emp\_status (or emp\_status by pov\_status) counts were not available in the macro data. 

On the other hand, the Gaussian Copula method was able to capture this relationship and returned an association value (0.3) that is much closer to the expected value than what the conditional probabilities method returned. However, as shown in the figure, the overall dependency structure returned by Copula does not exactly match the expected dependency structure because this method estimates associations based on information from multiple geographical locations and does not consider location-specific bi-variate relationships. This is one of the limitations of copula-based approaches like SynC. The proposed approach (GenSyn), combines both copula and conditional probabilities and generates data with a dependency structure that is the closest to the expected one. 

\subsection{TAE vs Population Size}
\begin{figure}[!t]
\centering
 \includegraphics[width=0.9\linewidth]{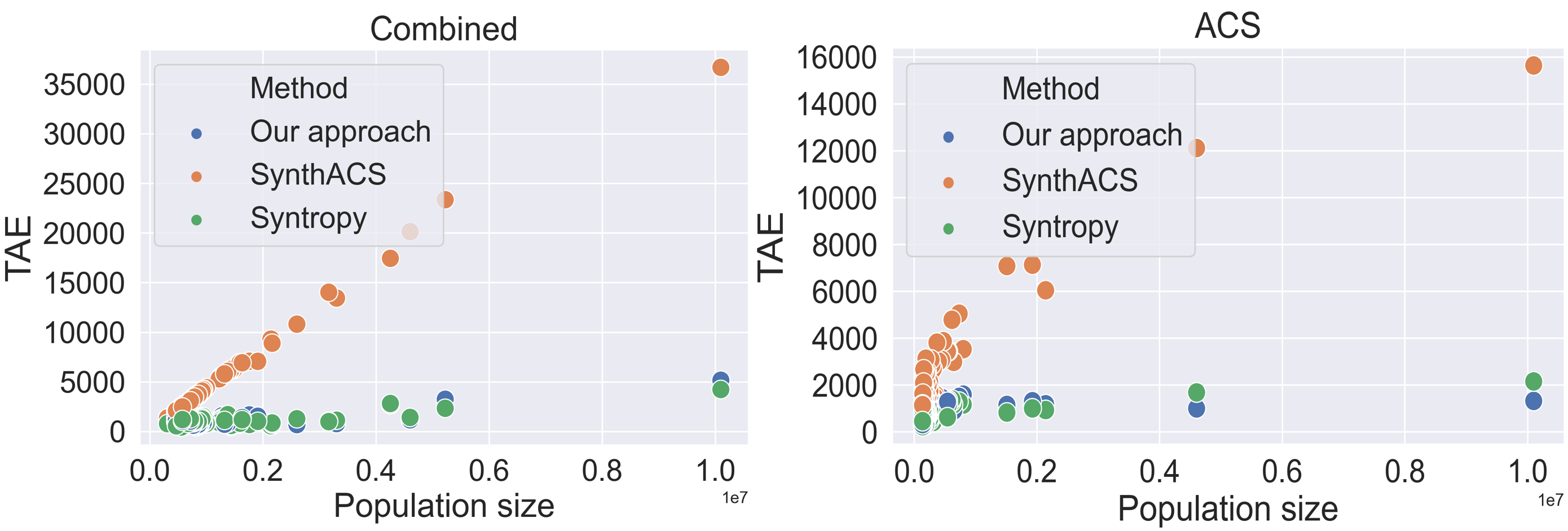}
\caption{Visualization of the relationship between TAE and population size for the three best performing methods}
\label{fig:pop_tAE}
\end{figure}

Fig. \ref{fig:measures} shows that the TAE values of GenSyn remain fairly low (less than 2000) for almost all the counties. However, it does not provide a good representation of how the performance changes with respect to population size. We visualize that with a scatterplot of the TAE values returned by GenSyn for the 50 counties (with varying population sizes) and compare it with two other best-performing methods (Syntropy and SynthACS). We do not report on SynC here because it returns 0 TAE values for any population size. As shown in Fig. \ref{fig:pop_tAE}, the TAE values for both GenSyn and Syntropy remain consistent for all population sizes which makes them suitable for all population sizes. On the contrary, the TAE value for SynthACS increases as the population size increases making it suitable only for small population sizes.

\begin{figure}[t]
    \centering
    \subfloat[\centering Overall performance]{{\includegraphics[width=0.9\linewidth]{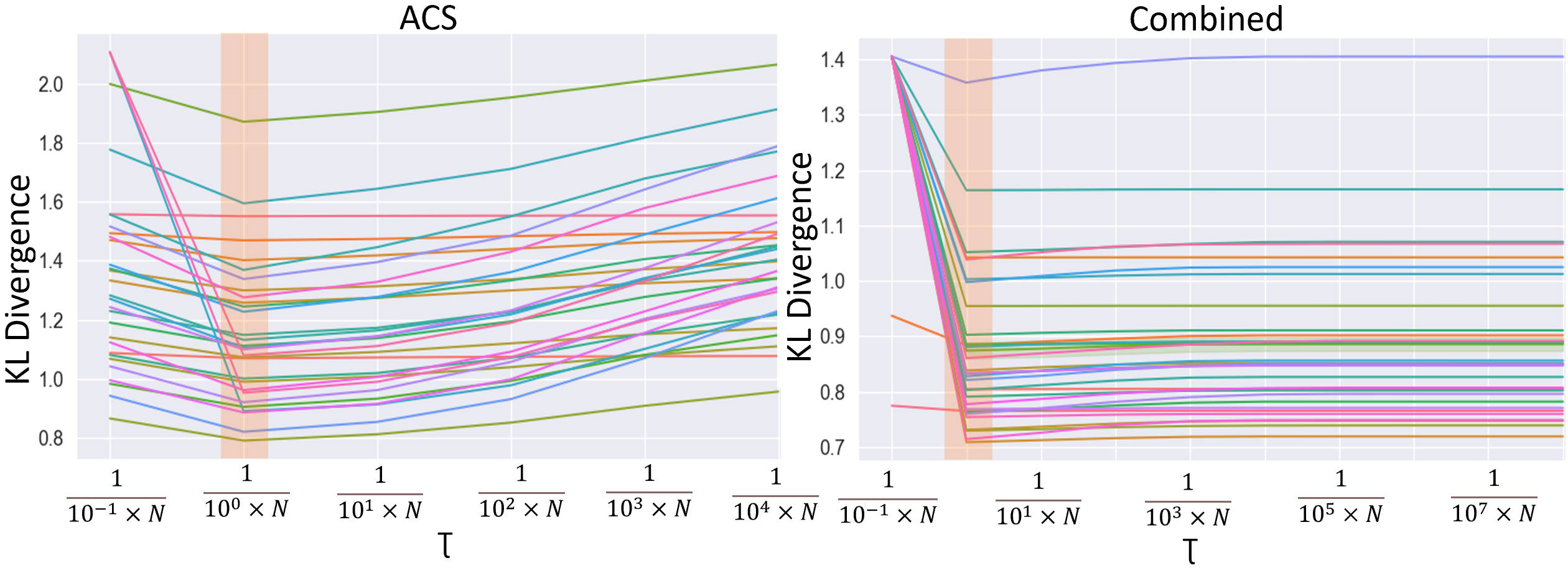}}}%
    \qquad
    \subfloat[\centering Performance on infrequent categories]{{\includegraphics[width=0.9\linewidth]{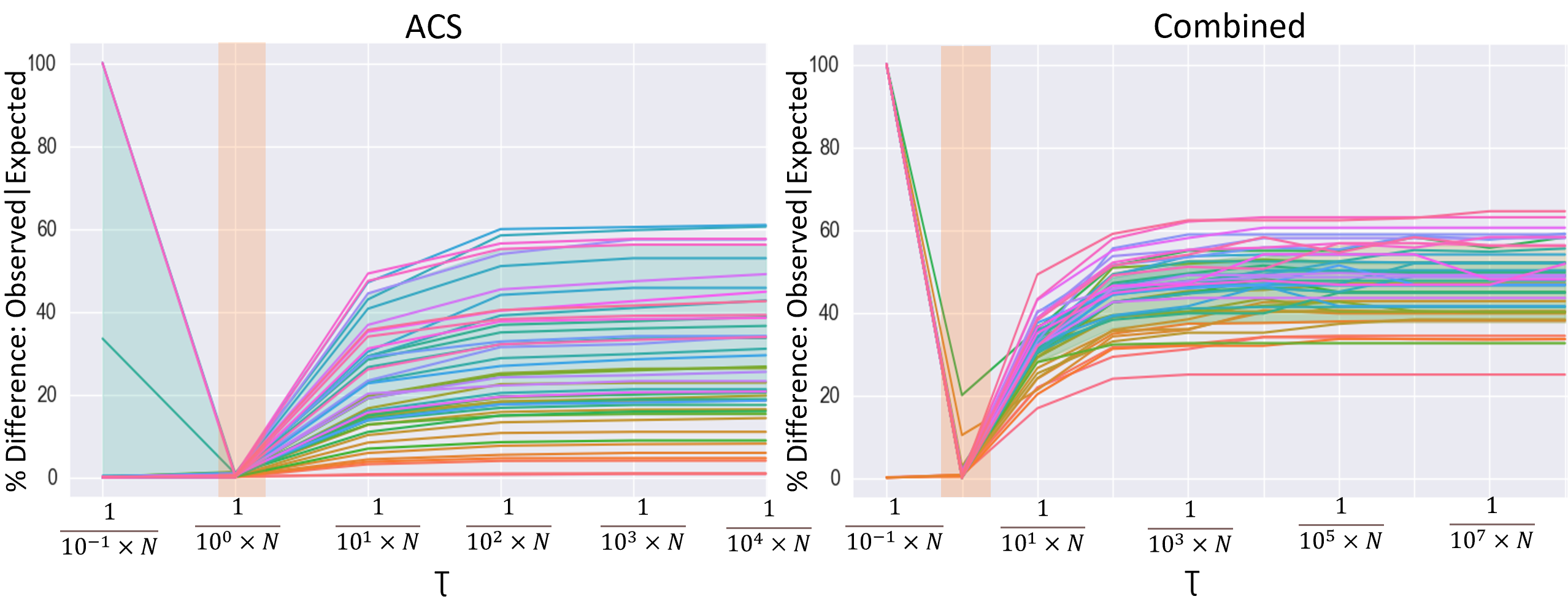} }}%
    \caption{Visualization of the relationship between $\tau$ and (a) the overall KL divergence, and (b) \% differences between the expected and observed frequency counts for infrequent categories. The plots present results for all counties (differentiated by colors) for both our datasets and highlights (in orange) the results for $\tau$ = $\frac{1}{N}$ which is the value we select}%
    \label{fig:tau}%
\end{figure}

 \begin{figure*}[!t]
\centering
 \includegraphics[width=0.65\linewidth]{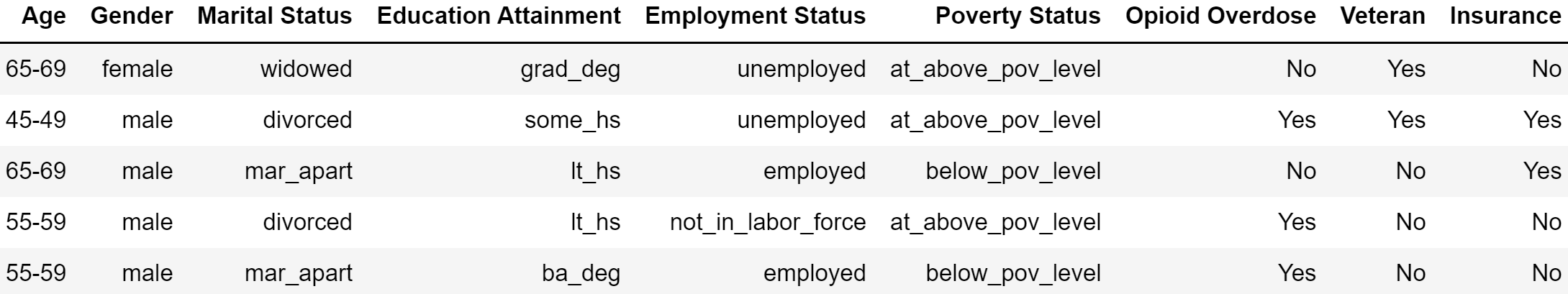}
\caption{A snippet of the generated synthetic population data}
\label{fig:synthetic-eg}
\end{figure*} 
\subsection{Sensitivity Analysis based on $\tau$}
Fig. \ref{fig:tau} shows the findings from the experiments with different values of $\tau$. Here, the differences in the colors represent the different counties. As shown, the best performance is obtained when the value of $\tau$ is set to $\frac{1}{N}$ (i.e., when we select profiles that have probabilities of occurring at least once in the synthetic population). As expected, the performance was poor at $\tau$ =  $\frac{1}{N \times 10^{-1}}$. It is possible that a lot of profiles that are present in the population got omitted when we filtered profiles with probabilities less than $\frac{1}{N \times 10^{-1}}$. The KL divergence values are very high for that $\tau$ (see Fig. \ref{fig:tau}(a)). To make the plots meaningful, extreme values of KL divergence were replaced with the maximum value of the non-extreme points. 

Fig. \ref{fig:tau}(b) shows how the change in $\tau$ impacts the performance in infrequent categories. Specifically, it shows the percentage difference between the expected and observed frequencies for the infrequent categories (opioid overdose and moved from abroad) against $\tau$. At $\tau$ =  $\frac{1}{N \times 10^{-1}}$, many counties have a percentage difference of 100 which implies that the model does not produce any profile that had an overdose or a profile that moved from abroad. It might be because infrequent categories are usually included in tuples with lower probabilities values which might have been filtered in the thresholding process. The overall performance and the performance on infrequent categories on the ACS dataset both decreased as the value of $\tau$ decreased (increase in the number of tuple spaces to be explored). A possible explanation for this is that the number of unknown parameters in the model increases as the number of population profiles grows, making the optimization with maximum entropy harder. 
For the Combined dataset, the KL divergence values for $\tau \leq \frac{1}{N}$ are almost similar. However, the performance in the infrequent category decreases as the value of $\tau$ decreases.
Based on these findings, $\tau$ was selected to be equal to $\frac{1}{N}$.

\subsection{Generated Data}
An example of the generated synthetic data for the Combined dataset is shown in Fig. \ref{fig:synthetic-eg}. The data consists of categorical attributes from multiple domains which are useful for opioid overdose-related research. The proposed approach is expected to enable timely access to data across multiple geographical locations to facilitate rapid decision-making and help uncover trends in opioid overdose and their underlying risk factors to support public and private health, justice, and social services agencies in addressing unmet needs. The same concept can be used for a variety of other applications.

\section{Conclusion}
In this study, we introduced a novel multi-stage framework, GenSyn, for generating synthetic microdata using macro data sources. By performing extensive experiments on two real-world datasets (representing both generic and domain-specific applications), we demonstrated that the proposed approach is capable of reconstructing the desired population data for any kind of problem which has the required macro data available.

\bibliography{sample-base-cp.bib}{}
\bibliographystyle{IEEEtran}

\section{APPENDIX}
\subsection{Copula sampling - Matching marginals}
Algorithm 2 (Line 14) returns the initial sampled data (Y) of the N profiles in $S_c$, which is obtained using Gaussian copula sampling. However, at this point, the sampled individuals do not agree with the marginals of the macro data. Algorithm \ref{alg3}, which is motivated by \cite{li2020sync}, tackles this problem and returns a probability distribution that is consistent with the input macro data. For each categorical variable $X_k$, Algorithm \ref{Algo2} returns the probability distribution of the $c_k$ categories associated with that variable, for the N profiles. Thus, each variable can be modeled as a multinomial distribution Multinomial$(c_k, p_{ik})$, where $p_{ik}$ represents the probabilities associated with a profile i and categories $c_k$. At every step, we generate a random category c from the probability distribution $p_{ik}$ and check if the marginal total $\eta_k[c]$ is greater than 0 (i.e., if the marginal total for that category has been met). If $\eta_k[c]$ is still greater than 0, we select that choice and subtract 1 from $\eta_k[c]$. Else, a different choice is selected from the distribution and the $\eta_k$ for that category is subtracted by 1. We continue this until all the marginal totals have been met. Once this process is complete, we obtain the probability distribution $p(X_1, X2,..., X_K)$ of $S_c$ based on the profiles returned from this step.
\begin{algorithm}[h]
\caption{Copula sampling - Matching marginals}
\label{alg3}
\KwIn{Initial sampled data (Y) from Algorithm 2, Marginal vector $\eta$ for $G_l$}
\KwOut{Joint probability distribution $p(X_1,X2,...,X_K)$ for the tuple space $S_c$}
K $\gets$ Number of variables\\
$c_k \gets$  categories of variable $X_k$\\
\For {i in N}{
\For{ k in K}{
$p_{ik} = Y[i, c_k]$\\
Draw a class $c$ from Multinomial$(c_k, p_{ik})$\\
\uIf{$\eta_k[c] >0$}{
$Y^{'}_{ik} = c$\\
$\eta_k[c] = \eta_k[c]-1$\\
}
\Else{goto 6}
}
$Y^{'}_i \gets [Y^{'}_{i1},....Y^{'}_{iK}]$
}
$Y^{'} \gets [Y^{'}_1,....Y^{'}_N]$\\
Get the probability distribution $p(X_1,..X_{i-1}, X_K)$ of tuple space $S_c$ based on Y'\\
\Return {$p(X_1,..X_{i-1}, X_K)$}
\end{algorithm}

\subsection{Modeling dependencies}
\begin{figure}[!htb]
\centering
 \includegraphics[width=0.45\linewidth]{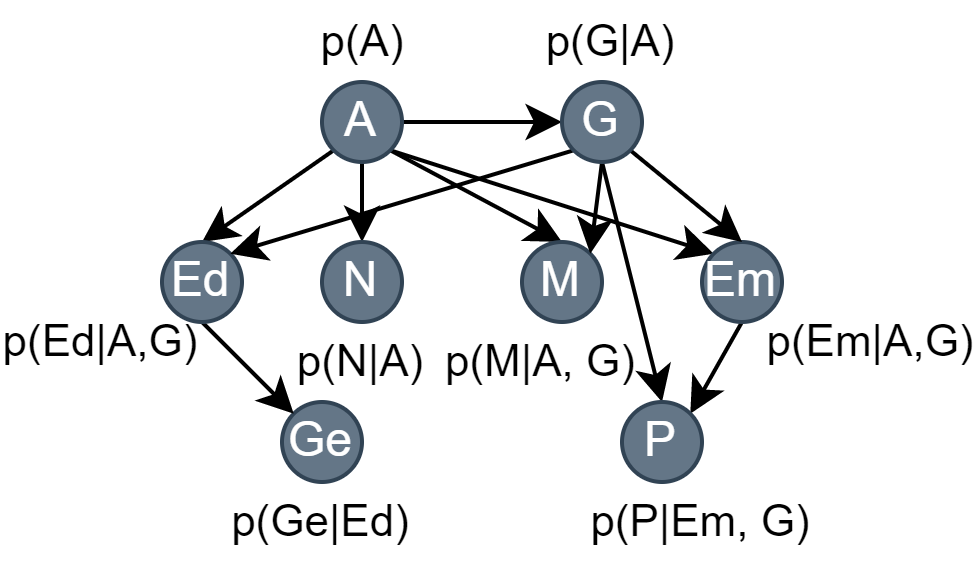}
\caption{Dependency graph representing macro data D2 for the ACS variables}
\label{fig:dep}
\end{figure}

Fig. \ref{fig:dep} depicts the dependency graph created for the ACS variables using the conditional relationships specified in Table \ref{tab:data}. As shown, each node represents different variables in the dataset and the arrows represent the variable conditioning as defined in the table. Once the dependency graph is created, we can traverse through the nodes in the graph based on which level they are in and define an ordering to be used in the conditional probabilities method. In this case, age (A) and gender (G) clearly have to be modeled first because other variables are conditioned on them. This is followed by the nodes in the second level (marital status (M), education attainment (Em), employment status (Ed), and nativity (N)) and then the third one (poverty status (P) and geographic mobility (Ge)). Within each level, the variable ordering can either be random or based on their entropy values. SynthACS \cite{synthacs} states that the best result is obtained when low entropy variables are added first followed by higher entropy ones. We followed SynthACS' hierarchy for ordering the ACS variables. 

For the Combined dataset, which also consists of variables from sources other than ACS (i.e., Opioid overdose, Insurance, and Veteran), ACS attributes were selected first and the rest of the attributes were randomly ordered because the results did not change significantly as the ordering changed.

\end{document}